\documentclass{article}
\usepackage[accepted]{icml2019}

\usepackage[T1]{fontenc}
\usepackage{tgtermes}
\usepackage{amssymb}
\newcommand{\mathbold}[1]{\ensuremath{\boldsymbol{\mathbf{#1}}}}
\usepackage[ttdefault=true]{AnonymousPro}

\usepackage{amsfonts}
\usepackage{amsmath}
\usepackage{amsthm}
\usepackage{nicefrac}
\usepackage{bm}

\usepackage[english]{babel}
\usepackage[parfill]{parskip}
\usepackage{afterpage}
\usepackage{enumitem}
\usepackage{framed}
\usepackage{xspace}

\usepackage{lineno}

\newcounter{parcount}

\usepackage[usenames,dvipsnames]{xcolor}
\definecolor{shadecolor}{gray}{0.9}

\usepackage{graphicx}
\usepackage[labelfont=bf]{caption}
\usepackage{subcaption}

\usepackage{booktabs, array}

\usepackage{listings}
\usepackage{fancyvrb}
\fvset{fontsize=\small}

\usepackage{natbib}
\usepackage[colorlinks,linktoc=all]{hyperref}
\usepackage[all]{hypcap}
\hypersetup{citecolor=MidnightBlue}
\hypersetup{linkcolor=MidnightBlue}
\hypersetup{urlcolor=MidnightBlue}

\usepackage[nameinlink]{cleveref}
\creflabelformat{equation}{#2\textup{#1}#3}  

\usepackage
[acronym,smallcaps,nowarn,section,nonumberlist]{glossaries}
\glsdisablehyper{}

\definecolor{strings}{rgb}{.624,.251,.259}
\definecolor{keywords}{rgb}{.224,.451,.686}
\definecolor{comment}{rgb}{.322,.451,.322}

\lstdefinelanguage{python}{
  morekeywords={from, import, as, for, in, while, def, return, =, +, if, elif, else, with,
  -, /, *, lambda, global, del},
  keywords=[2]{1,2,3,4,5,6,7,8,9,0, __init__, STACK, class, super, self},
  keywords=[3]{list,len},
  morecomment=[l]{\#},
  morecomment=[s]{"""}{"""},
  morestring=[b]',
  morestring=[b]",
  alsoletter={<>=-+/*},
  sensitive=true
}

\renewcommand{\d}[1]{\ensuremath{\operatorname{d}\!{#1}}}
\newcommand{\nestedmathbold}[1]{{\mathbold{#1}}}

\newcommand{\mbb}{\nestedmathbold{b}}

\newcommand{\mbh}{\nestedmathbold{h}}

\newcommand{\mbx}{\nestedmathbold{x}}
\newcommand{\mby}{\nestedmathbold{y}}

\newcommand{\mbW}{\nestedmathbold{W}}

\usepackage{tikz}

\usetikzlibrary{bayesnet} 

\pgfdeclarelayer{edgelayer}
\pgfdeclarelayer{nodelayer}
\pgfsetlayers{edgelayer,nodelayer,main}

\definecolor{hexcolor0xbfbfbf}{rgb}{0.749,0.749,0.749}

\tikzset{>=latex}
\tikzstyle{none}   = [inner sep=0pt]
\tikzstyle{line}   = [ -, thick, shorten <=1pt, shorten >=1pt ]
\tikzstyle{arrow}  = [ ->, thick, shorten <=1pt, shorten >=1pt ]
\tikzstyle{ardash} = [ dashed, ->, thick, shorten <=1pt, shorten >=1pt ]

\tikzstyle{empty}=[circle,opacity=0.0,text opacity=1.0,inner sep=0pt]
\tikzstyle{box}=[rectangle,fill=White,draw=Black]
\tikzstyle{filled}=[circle,thick,fill=hexcolor0xbfbfbf,draw=Black]
\tikzstyle{hollow}=[circle,thick,fill=White,draw=Black]
\tikzstyle{param}=[rectangle,fill=Black,draw=Black,inner sep=0pt,minimum width=4pt,minimum height=4pt]
\tikzstyle{paramhollow}=[rectangle,thick,fill=White,draw=Black,inner sep=0pt,minimum
width=4pt,minimum height=4pt]

\usepackage{pgfplots}                               
\pgfplotsset{compat=newest}
\pgfplotsset{plot coordinates/math parser=false}
\newlength\figureheight
\newlength\figurewidth
\newlength\figureheightsmall
\newlength\figurewidthsmall
\definecolor{POSTcolor}{rgb}{0.48, 0.20, 0.58} 
\definecolor{Qcolor}{rgb}{0.00, 0.53, 0.22} 

\input{_minted-main/default-pyg-prefix.pygstyle}
\input{_minted-main/default.pygstyle}

\usepackage{ragged2e}

\DeclareFontShape{OT1}{cmtt}{bx}{n}{<5><6><7><8><9><10><10.95><12><14.4><17.28><20.74><24.88>cmttb10}{}

\icmltitlerunning{Bayesian Layers: A Module for Neural Network Uncertainty}

\begin{document}

\title{Bayesian Layers: A Module for \\ Neural Network Uncertainty}

\twocolumn[
\icmltitle{Bayesian Layers: A Module for Neural Network Uncertainty}

\icmlsetsymbol{equal}{*}

\begin{icmlauthorlist}
\icmlauthor{Dustin Tran}{google}
\icmlauthor{Michael W. Dusenberry}{google}
\icmlauthor{Mark van der Wilk}{prowler}
\icmlauthor{Danijar Hafner}{google}
\end{icmlauthorlist}

\icmlaffiliation{google}{Google Brain, Mountain View, California, USA}
\icmlaffiliation{prowler}{Prowler.io, London, United Kingdom}

\icmlcorrespondingauthor{Dustin Tran}{trandustin@google.com}

\icmlkeywords{Machine Learning, ICML}

\vskip 0.3in
]

\printAffiliationsAndNotice{}

\begin{abstract}
\vskip 0.1in
We describe Bayesian Layers, a module designed for fast experimentation with neural network uncertainty. It extends neural network libraries with drop-in replacements for common layers. This enables composition via a unified abstraction over deterministic and stochastic functions and allows for scalability via the underlying system. These layers capture uncertainty over weights (Bayesian neural nets), pre-activation units (dropout), activations (``stochastic output layers''), or the function itself (Gaussian processes). They can also be reversible to propagate uncertainty from input to output. We include code examples for common architectures such as Bayesian LSTMs, deep GPs, and flow-based models. As demonstration, we fit a 5-billion parameter ``Bayesian Transformer'' on 512 TPUv2 cores for uncertainty in machine translation and a Bayesian dynamics model for model-based planning.
Finally, we show how Bayesian Layers can be used within the Edward2 probabilistic programming language for probabilistic programs with stochastic processes.
  \footnote{All code is available at \url{https://github.com/tensorflow/tensor2tensor}. Dependency-wise, it extends Keras in TensorFlow \citep{chollet2015keras} and uses Edward2 \citep{tran2018simple} to operate with random variables.
Namespaces: \texttt{import tensorflow as tf; ed=edward2}. Code snippets assume \texttt{tensorflow==1.12.0}.}
\end{abstract}

\begin{figure}[!htb]
\centering
\begin{minipage}{\columnwidth}
  \centering
  \input{_minted-main/64C054396F9D630507A12505E63F104EB722149D8701C8F8F9BC8FCEDA70B7B8.pygtex}

\captionof{figure}{Bayesian RNN \citep{fortunato2017bayesian}. Bayesian Layers integrate easily into existing workflows.
}
\label{fig:bnn}
\end{minipage}%
\hfill
\begin{minipage}{\columnwidth}
  \centering
  \centering
  \begin{tikzpicture}

  \node[empty]              (dot)      {} ;
  \node[obs, right=of dot, xshift=-0.75cm]        (xt) {$\mbx_t$} ;
  \node[latent, left=of xt, xshift=0.75cm, yshift=-0.75cm]  (bh)  {$\mbb_h$} ;
  \node[latent, above=of bh, yshift=-0.9cm]      (Wx) {$\mbW_x$} ;
  \node[latent, above=of Wx, yshift=-0.9cm, xshift=0.1cm]      (Wh)      {$\mbW_h$} ;
  
  \factor[below=1.2cm of bh, yshift=0.9cm] {Wy} {} {} {};
  \node[left=-0.05cm of Wy, yshift=-0.0cm] (Wylabel) {$\mbW_y$};
  \factor[below=1.4cm of Wy, yshift=0.9cm] {by} {} {} {};
  \node[left=-0.05cm of by, yshift=-0.0cm] (bylabel) {$\mbb_y$};

  \factor[below=0.75cm of xt] {ht} {} {} {};
  \node[right=0.03cm of ht, yshift=-0.3cm] (htlabel) {$\mbh_t$};
  \node[empty, left=of ht]      (htminus)    {$\cdots$} ;
  \node[empty, right=of ht]      (htplus)    {$\cdots$} ;

  \node[obs, below=0.75cm of ht]      (yt)    {$\mby_t$} ;
  \node[empty, left=of yt]      (ytminus)    {} ;
  \node[empty, right=of yt]      (ytplus)    {} ;

  \edge{Wh}{ht};
  \edge{Wx}{ht};
  \edge{bh}{ht};
  \edge{xt}{ht};
  \edge{Wy}{yt};
  \edge{by}{yt};
  \edge{ht}{yt};
  \edge{htminus}{ht};
  \edge{ht}{htplus};

\end{tikzpicture}
\captionof{figure}{
Graphical model depiction. Default arguments specify learnable distributions over the LSTM's weights and biases; we apply a deterministic output layer.}
\label{fig:bnn-graphical}
\end{minipage}
\vspace{-3ex}
\end{figure}

\section{Introduction}
\label{sec:intro}

The rise of AI accelerators such as TPUs lets us utilize computation with $10^{16}$ FLOP/s and 4 TB of memory distributed across hundreds of processors \citep{jouppi2017datacenter}. In principle, this lets us fit probabilistic models at
many orders of magnitude larger than state of the art.
We are particularly inspired by research on uncertainty-aware functions: priors and algorithms for Bayesian neural networks \citep[e.g., ][]{wen2018flipout,hafner2018reliable}, scaling up Gaussian processes \citep[e.g., ][]{salimbeni2017doubly,john2018large}, and
expressive distributions via invertible functions \citep[e.g., ][]{rezende2015variational}.

Unfortunately, while research with uncertainty-aware functions are not limited by hardware, they are limited by software.
Modern systems approach this by inventing a probabilistic programming language which encompasses all computable probability models as well as a universal inference engine \citep{goodman2012church,carpenter2016stan} or with composable inference \citep{tran2016edward,bingham2018pyro,probtorch2017probtorch}. Alternatively, the software may use high-level abstractions in order to specify and fit specific model classes with a hand-derived algorithm \citep{gpy2014,vanhatalo2013gpstuff,GPflow2017}. These systems have all met success, but they tend to be monolothic in design. This prevents research flexibility such as utilizing low-level communication primitives to truly scale up models to billions of parameters.

Most recently, Edward2 provides lower-level flexibility by enabling arbitrary numerical ops with random variables
\citep{tran2018simple}. However, it remains unclear how to leverage random variables for uncertainty-aware functions. For example, current practices with Bayesian neural networks require explicit network computation and variable management \citep{tran2016edward} or
require indirection by intercepting
weight instantiations of a deterministic layer \citep{bingham2018pyro}.
Both designs are inflexible for many real-world uses in research. In practice, researchers often use the lower numerical level---without a unified design for uncertainty-aware functions as there are for deterministic neural networks and automatic differentiation. This forces researchers to reimplement even basic methods such as Bayes by Backprop \citep{blundell2015weight}---let
alone build on and scale up more complex baselines.

This paper describes Bayesian Layers, an extension of neural network libraries which contributes one idea: instead of only deterministic functions as ``layers'', enable distributions over functions.
Bayesian Layers does not invent a new language. It inherits neural network semantics to specify uncertainty-aware models as a composition of layers.
Each layer may capture uncertainty over weights (Bayesian neural nets), pre-activation units (dropout), activations (``stochastic output layers''), or the function itself (Gaussian processes). They can also be reversible layers that propagate uncertainty from input to output.

We include code examples for common architectures such as Bayesian LSTMs, deep GPs, and flow-based models. As demonstration, we fit a 5-billion parameter ``Bayesian Transformer'' on 512 TPUv2 cores for uncertainty in machine translation and a Bayesian dynamics model for model-based planning. Finally, we show how Bayesian Layers can be used as primitives in the Edward2 probabilistic programming language.

\subsection{Related Work}
\label{sub:related}

There have been many software developments for distributions over functions. Our work takes classic inspiration from Radford Neal's software in 1995 to enable flexible modeling with both Bayesian neural nets and GPs \citep{neal1995software}.
Modern software typically focuses on only one of these directions.
For Bayesian neural nets, researchers have commonly coupled variational sampling in neural net layers (e.g., code and algorithms from \citet{gal2016dropout,louizos2017multiplicative}).
For Gaussian processes, there have been significant developments in libraries \citep{rasmussen2010gaussian,gpy2014,vanhatalo2013gpstuff,GPflow2017,al2017learning,gardner2018gpytorch}.
Perhaps most similar to our work, Aboleth \citep{aboleth2017aboleth} features variational BNNs and random feature approximations for GPs.
Aside from API differences from all these works, our work tries to revive the spirit of enabling any function with uncertainty---whether that be, e.g., in the weights, activations, or the entire function---and to do so in a manner compatible with scalable deep learning ecosystems.

A related thread are probabilistic programming languages which build on the semantics of an existing functional programming language. Examples include HANSEI on OCaml, Church on Lisp, and Hakaru on Haskell \citep{kiselyov2009embedded,goodman2012church,narayanan2016probabilistic}.
Neural network libraries can also be thought of as a (fairly simple) functional programming language, with limited higher-order logic
and a type system of (finite lists of) $n$-dimensional arrays.
Unlike the above probabilistic programming approaches, Bayesian Layers doesn't introduce new primitives to the underlying language. As we describe next, it overloads the existing primitives with a method to handle randomness in any state in its execution.

\section{Bayesian Layers}
In neural network libraries, architectures decompose as a composition of ``layer'' objects as the core building block \citep{collobert2011torch,alrfou2016theano,jia2014caffe,chollet2015keras,chen2015mxnet,tensorflow2015-whitepaper,guadarrama2016slim}. These layers capture both the parameters and computation of a mathematical function into a programmable class.

In our work, we extend layers to capture
``distributions over functions'', which we describe as a layer with uncertainty about some state in its computation---be it uncertainty in the weights, pre-activation units, activations, or the entire function. Each sample from the distribution instantiates a different function, e.g., a layer with a different weight configuration.

\subsection{Bayesian Neural Network Layers}
The Bayesian extension of any deterministic layer is to place a prior distribution over its weights and biases. These layers require several considerations.
\Cref{fig:bnn} implements a Bayesian RNN.

\paragraph{Computing the integral}
We need to compute often-intractable integrals over weights and biases $\theta$. Consider for example two cases, the variational objective for training and the approximate predictive distribution for testing,
\begin{align*}
\textrm{ELBO}(\theta) &= \int q(\theta) \log p(\mathbf{y}\mid f_\theta(\mathbf{x})) \d\theta \\ & \hspace{1em}- \mathrm{KL}\left[q(\theta)\,\|\,p(\theta)\right],
\\
q(\mathbf{y}\mid\mathbf{x}) &= \int q(\theta) p(\mathbf{y}\mid f_\theta(\mathbf{x})) \d\theta.
\end{align*}
Here, $\mathbf{x}$ may be a real-valued tensor as a batch of input features, $\mathbf{y}$ may be a vector-valued output for each feature set, and the function $f$ encompasses the overall network as a composition of layers.

To enable different methods to estimate
these integrals,
we implement each estimator as its own Layer.
The same Bayesian neural net can use entirely different computational graphs depending on the estimation (and therefore entirely different code).
For example, sampling from $q(\theta)$ with reparameterization and running the deterministic layer computation is a generic way to evaluate layer-wise integrals \citep{kingma2014auto}. Alternatively, given small weight dimensions, one could approximate each integral deterministically via quadrature.
To enable modularity, the only restriction is that the overall integral estimation can be decomposed layer-wise; \Cref{sec:limitations} describes such restrictions in more depth.

\begin{figure}[!tb]
  \input{_minted-main/76F0FCDC3BFC2031B3EA814CB868ECC0B722149D8701C8F8F9BC8FCEDA70B7B8.pygtex}
\caption{Bayesian layers are modularized to fit existing neural net semantics of initializers, regularizers, and layers as they deem fit. Here, a Bayesian layer with reparameterization \citep{kingma2014auto,blundell2015weight} is the same as its deterministic implementation. The only change is the default for \texttt{kernel\_\{initializer,regularizer\}}; no additional methods are added.
}
\label{fig:dense}
\end{figure}

\paragraph{Signature}
We'd like to have the Bayesian extension of a deterministic layer retain its mandatory
constructor arguments as well as its call signature of tensor-dimensional inputs and tensor-dimensional outputs.
This avoids cognitive overhead, letting one easily swap layers (\Cref{fig:swap}; \citet{laumann2018bayesian}).
For example, a dense (feedforward) layer requires a \texttt{units} argument determining its output dimensionality; a convolutional layer also includes \texttt{kernel\_size}.
We maintain optional arguments (and add new ones) if they make sense.

\begin{figure}[!tb]
  \input{_minted-main/8C5FC8604A34506818C04D99A2123998B722149D8701C8F8F9BC8FCEDA70B7B8.pygtex}

\caption{Bayesian Layers are drop-in replacements for their deterministic counterparts.
}
\label{fig:swap}
\end{figure}

\paragraph{Distributions over parameters} To specify distributions, a natural idea is to overload the existing parameter initialization arguments in a Layer's constructor; in Keras, it is
\texttt{kernel\_initializer} and \texttt{bias\_initializer}. These arguments are extended to accept callables that take metadata such as input shape and
return a distribution over the parameter. Distribution initializers may carry trainable parameters, each with their own initializers (pointwise or distributional). The default initializer represents a trainable approximate posterior in a variational inference scheme (\Cref{fig:dense}).

For the distribution abstraction, we use Edward \texttt{RandomVariables} \citep{tran2018simple}. They are Tensors augmented with distribution methods such as \texttt{sample} and \texttt{log\_prob}; by default, numerical ops operate on its sample Tensor.
Layers perform forward passes using deterministic ops and the \texttt{RandomVariables}.
\paragraph{Distribution regularizers} 
The variational training objective requires the evaluation of a KL term, which penalizes deviations of the learned $q(\theta)$ from the prior $p(\theta)$. Similar to distribution initializers, we overload the existing parameter regularization arguments in a layer's constructor; in Keras, it is
\texttt{kernel\_regularizer} and \texttt{bias\_regularizer} (\Cref{fig:dense}). These arguments are extended to accept callables that take in the kernel or bias \texttt{RandomVariables} and return a scalar Tensor. By default, we use a KL divergence toward the standard normal distribution, which represents the penalty term common in variational Bayesian neural network training.

Explicitly returning regularizers in a Layer's call ruins composability (see Signature above). Therefore Bayesian layers, like their deterministic counterparts, side-effect the computation: one queries an attribute to access any regularizers for, e.g., the loss function. \Cref{fig:bnn} implements a Bayesian RNN; \Cref{appendix:cnn} implements a Bayesian CNN (ResNet-50).

\subsection{Gaussian Process Layers}

As opposed to representing distributions over functions through the weights, Gaussian processes represent distributions over functions by specifying the value of the function at different inputs.
Recent advances have made Gaussian process inference computationally similar to Bayesian neural networks  \citep{hensman2013svi}. We only require a method to sample the function value at a new input, and evaluate KL regularizers. This allows GPs to be placed in the same framework as above.\footnote{%
More broadly, these ideas extend to stochastic processes. For example, we plan to implement a Poisson process layer for scalable point process modeling.
}
\Cref{fig:gp} implements a deep GP.

The considerations are the same as above, and we make similar decisions:

\paragraph{Computing the integral}
We use a separate class for each estimator. This includes \texttt{GaussianProcess} for exact integration, which is only possible in limited situations; \texttt{SparseGaussianProcess} for inducing variable approximations; and \texttt{RandomFourierFeatures} for projection approximations.

\paragraph{Signature}
For the equivalent deterministic layer, maintain its mandatory arguments as well as tensor-dimensional inputs and outputs. For example, the number of \texttt{units} in a Gaussian process layer determine the GP's output dimensionality, where \texttt{layers.GaussianProcess(32)} is the Bayesian nonparametric extension of \texttt{tf.keras.layers.Dense(32)}. Instead of an optional \texttt{activation} function argument, GP layers have optional mean and covariance function arguments which default to the zero function and squared exponential kernel respectively. We also include an optional argument for what set of inputs and outputs to condition on: this allows the GP layer to perform both prior and posterior predictive computations. Any state in the layer's computational graph may be trainable---whether they be kernel hyperparameters or the inputs and outputs that function conditions on.

\paragraph{Distribution regularizers}
By default, we include no regularizer for exact GPs, a KL divergence regularizer on the inducing output distribution for sparse GPs, and a KL divergence regularizer on weights for random projection approximations. These defaults reflect each inference method's standard for training, where the KL regularizers use the same implementation as the Bayesian neural nets'.

\begin{figure}[t]
  \input{_minted-main/7EE0937A188822155A93843306B92A35B722149D8701C8F8F9BC8FCEDA70B7B8.pygtex}

\caption{Three-layer deep GP with variational inference \citep{salimbeni2017doubly,damianou2013deep}. We apply it for regression given batches of spatial inputs and vector-valued outputs. We flatten inputs to use the default squared exponential kernel; this naturally extends to pass in a more sophisticated kernel function.
}
\label{fig:gp}
\end{figure}

\subsection{Stochastic Output Layers}

\begin{figure}[!tb]
\centering
  \input{_minted-main/40001B2C8F0FF591E983045B25796D7EB722149D8701C8F8F9BC8FCEDA70B7B8.pygtex}

\caption{Image Transformer with discretized logistic mixture \citep{parmar2018image} over 128x128x3 features. Stochastic output layers let one easily experiment with the likelihood. We assume layers which don't exist in Keras; functional versions are available in Tensor2Tensor \citep{vaswani2018tensor2tensor}.
}
\label{fig:img-transformer}
\end{figure}

\begin{figure}[!htb]
\centering
  \input{_minted-main/AC5B561AC16BABCFEC6BA35D02239BA2B722149D8701C8F8F9BC8FCEDA70B7B8.pygtex}

\caption{A variational auto-encoder for compressing 256x256x3 ImageNet into a 32x32x3 latent code. Stochastic output layers are a natural approach for specifying stochastic encoders and decoders, and utilizing their log-probability or KL divergence.
}
\label{fig:vae}
\end{figure}

In addition to uncertainty over the \emph{mapping} defined by a layer, we may want to simply add stochasticity to the output. These outputs have a tractable distribution, and we often would like to access its properties: for example, maximum likelihood with an autoregressive network whose output is a discretized logistic mixture \citep{salimans2017pixelcnnpp}  (\Cref{fig:img-transformer}); or an auto-encoder with stochastic encoders and decoders (\Cref{fig:vae}).%
\footnote{
In previous figures, we used loss functions such as \texttt{mean\_squared\_error}. With stochastic output layers, we can replace them with a layer returning the likelihood and calling \texttt{log\_prob}.
}

\paragraph{Signature}
To implement stochastic output layers, we perform deterministic computations given a tensor-dimensional input and return a \texttt{RandomVariable}. Because \texttt{RandomVariables} are Tensor-like objects, one can operate on them as if they were Tensors: composing stochastic output layers is valid. In addition, using such a layer as the last one in a network allows one to compute properties such as a network's entropy or likelihood given data.

Stochastic output layers typically don't have mandatory constructor arguments. An optional \texttt{units} argument determines its output dimensionality (operated on via a trainable linear projection); the default maintains the input shape and has no such projection.

\subsection{Reversible Layers}
\label{sub:reversible}

\begin{figure}[t]
\input{_minted-main/4B490A3109DF97F9E67D9E88118374EAB722149D8701C8F8F9BC8FCEDA70B7B8.pygtex}

\caption{A flow-based model for image generation \citep{dinh2017density}.
}
\label{fig:reversible}
\end{figure}

With random variables in layers, one can naturally capture invertible neural networks which propagate uncertainty from input to output. This allows one to perform transformations of random variables, ranging from simple transformations such as for a log-normal distribution or high-dimensional transformations for flow-based models.

We make two considerations to design reversible layers:

\paragraph{Inversion}
Invertible neural networks are not possible with current libraries.
A natural idea is to design a new abstraction for invertible functions \citep{dillon2017tensorflow}. Unfortunately, this prevents interoperability with existing layer and model abstractions. Instead, we simply overload the notion of a ``layer'' by adding an additional method \texttt{reverse} which performs the inverse computation of its call and optionally \texttt{log\_det\_jacobian}. A higher-order layer called \texttt{layers.Reverse} takes a layer as input and returns another layer swapping the forward and reverse computation; by ducktyping, the reverse layer raises an error only during its call if \texttt{reverse} is not implemented. Avoiding a new abstraction both simplifies usage and also makes reversible layers compatible with other higher-order layers such as \texttt{tf.keras.Sequential}, which returns a composition of a sequence of layers.

\paragraph{Propagating Uncertainty}
As with other deterministic layers, reversible layers take a tensor-dimensional input and return a tensor-dimensional output where the output dimensionality is determined by its arguments. In order to propagate uncertainty from input to output, reversible layers may also take a \texttt{RandomVariable} as input and return a transformed \texttt{RandomVariable} determined by its call, \texttt{reverse}, and \texttt{log\_det\_jacobian}.%
\footnote{We implement \texttt{layers.Discretize} this way in \Cref{fig:img-transformer}. It takes a continuous \texttt{RandomVariable} as input and returns a transformed variable with probabilities integrated over bins.
}
Figure \ref{fig:reversible} implements RealNVP \citep{dinh2017density}, which is a reversible layer parameterized by another network (here, MADE \citep{germain2015made}). These ideas also extend to
reversible networks that enable backpropagation without storing intermediate activations in memory during the forward pass \citep{gomez2017reversible}.

\subsection{Layers for Probabilistic Programming}

\begin{figure}[t]
  \input{_minted-main/56A6FC1BDE99B9B03CF91B39CE5B09FBB722149D8701C8F8F9BC8FCEDA70B7B8.pygtex}

\caption{Cox process with a deep GP prior and a sparse GP posterior approximation. Using Bayesian Layers in a probabilistic programming language allows for a clean distinction between modeling and inference, as well as more flexible inference algorithms.
}
\label{fig:ppl}
\end{figure}

While the framework we laid out so far tightly integrates deep Bayesian modelling into existing ecosystems, we have deliberately limited our scope.
In particular, our layers tie the model specification to the inference algorithm (typically, variational inference). Bayesian Layers' core assumption is the modularization of inference per layer. This makes inference procedures which depend on the full parameter space, such as Markov chain Monte Carlo, difficult to fit within the framework.

\Cref{fig:ppl} shows how one can utilize Bayesian Layers in the Edward2 probabilistic programming language for more flexible modeling and inference. We could use, e.g., expectation propagation \citep{bui2016deepgpep,miguel2015pbp}. See \citet{tran2018simple} for details on how to use Edward2's tracing mechanisms for arbitrary training.

\section{Experiments}

We described a design for uncertainty-aware models built on top of neural network libraries. In experiments, we illustrate two points: 1. Bayesian Layers is efficient and makes possible new model classes that haven't been tried before (in either scale or flexibility); and 2. utilizing such Bayesian models provides benefits in applications including model-based planning.

\subsection{Model-Parallel Bayesian Transformer for \\\hspace{1.5em} Machine Translation}
\label{sub:transformer}

\begin{figure}[!htb]
\centering
\includegraphics[width=0.5\textwidth]{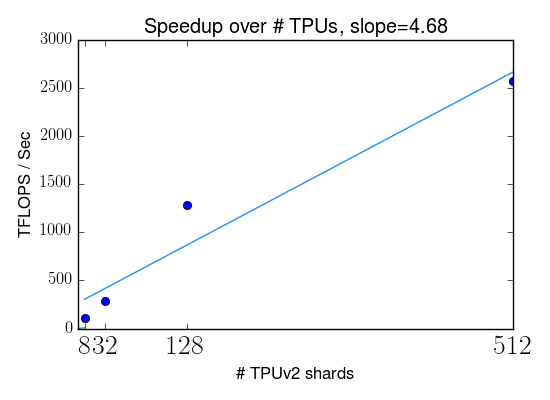}
\caption{Bayesian Transformer implemented with model parallelism ranging from 8 TPUv2 shards (core) to 512. As desired, the model's training performance scales linearly as the number of cores increases.
}
\label{fig:transformer}
\end{figure}

We implemented a ``Bayesian Transformer'' for the WMT14 EN-FR translation task. Using Mesh TensorFlow \citep{shazeer2018mesh}, we took a 2.8 billion parameter Transformer which reports a state-of-the-art BLEU score of 43.9. We then augmented the model with priors over the projection matrices by replacing calls to a multihead-attention layer with its Bayesian counterpart (using the Flipout estimator); we also made the pointwise feedforward layers Bayesian. \Cref{fig:transformer} shows that we can fit models with over 5-billion parameters (roughly twice as many due to a mean and standard deviation parameter), utilizing up to 2500 TFLOPs on 512 TPUv2 cores.

In attempting these scales, we were able to reach state-of-the-art perplexities
while achieving a higher predictive variance. This may suggest the Bayesian Transformr more correctly accounts for uncertainty given that the dataset is actually fairly small given the size of the model.
We also implemented a ``Bayesian Transformer'' for the One-Billion-Word Language Modeling Benchmark \citep{chelba2013one}, maintaining the same state-of-the-art perplexity of 23.1. We identified a number of challenges in both scaling up Bayesian neural nets and understanding their text applications; we leave this for future work separate from this systems paper.

\subsection{Bayesian Dynamics Model for Model-Based Reinforcement Learning}
\label{sub:rl}

\begin{figure*}[!t]
\newbox{\bigpicturebox}
\sbox{\bigpicturebox}{%
  \scalebox{1}[1.2]{\includegraphics[width=.72\textwidth]{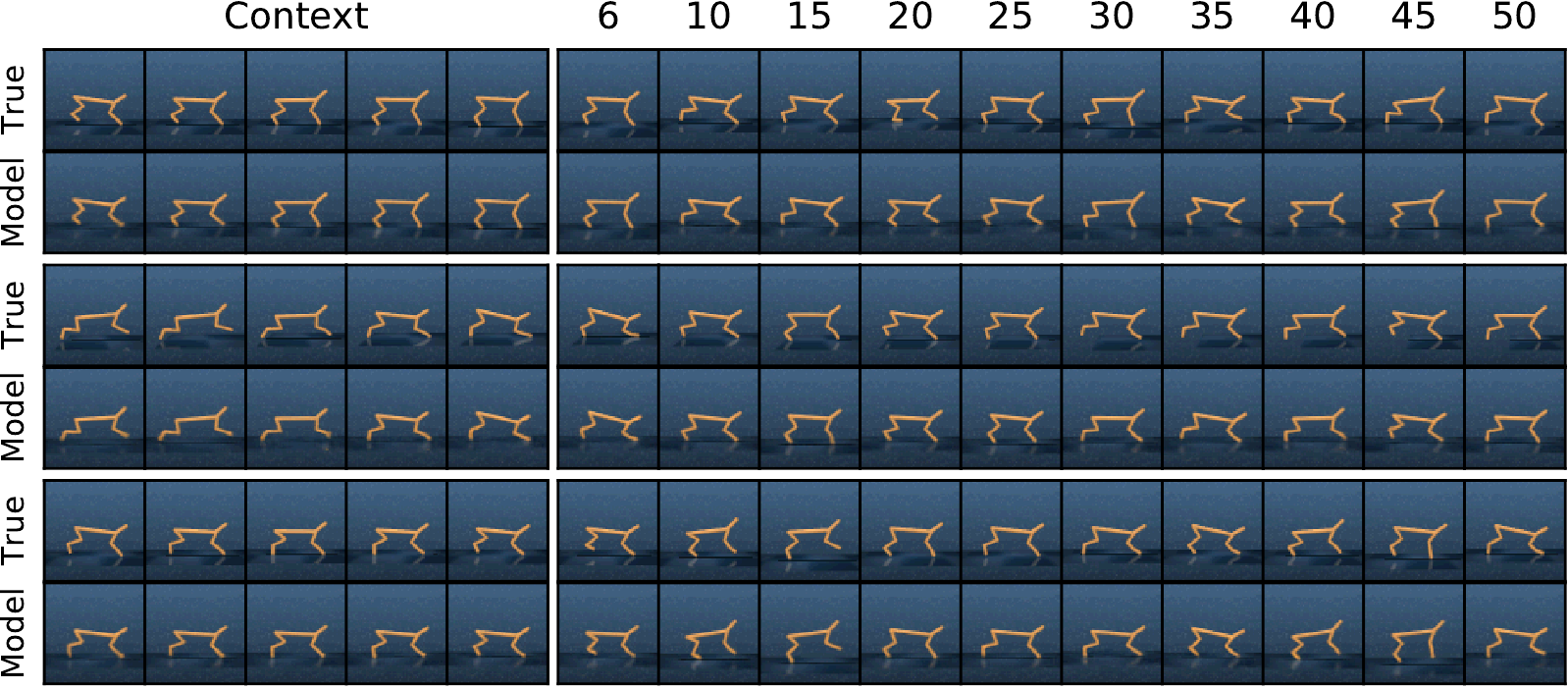}}%
}
\usebox{\bigpicturebox}\hfill
\begin{minipage}[b][\ht\bigpicturebox][s]{.25\textwidth}
\includegraphics[width=\textwidth]{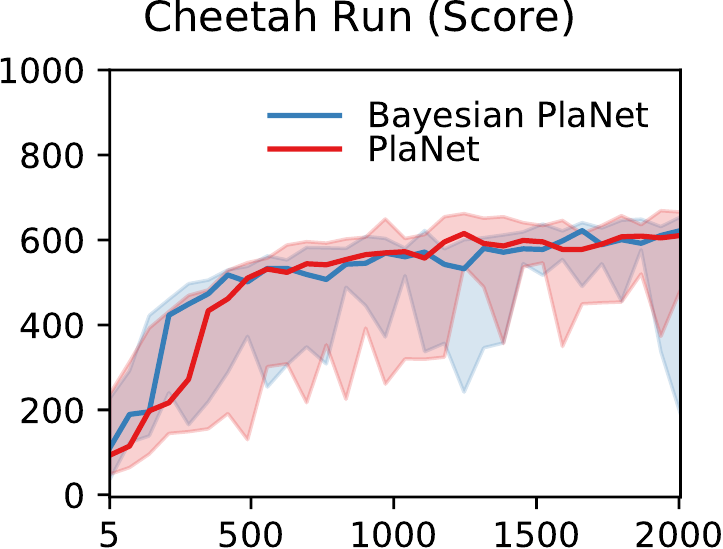}\\[2ex]
\includegraphics[width=\textwidth]{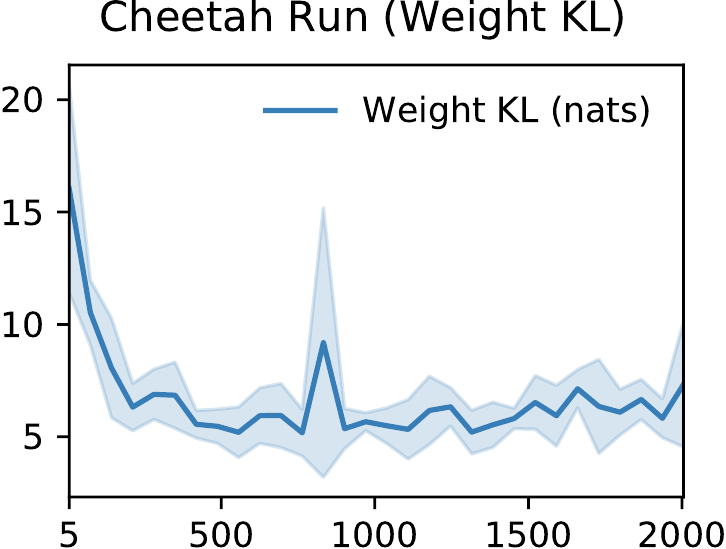}
\end{minipage}
\caption{Results of the Bayesian PlaNet agent. The score shows the task median performance over 5 seeds and 10 episodes each, with percentiles 5 to 95 shaded. Our Bayesian version of the method reaches the same task performance. The graph of the weight KL shows that the weight posterior learns a non-trivial function. The open-loop video predictions show that the agent can accurately make predictions into the future for 50 time steps.}
\label{fig:bnn_planet}
\end{figure*}

In reinforcement learning, uncertainty estimates can allow for directed exploration, safe exploration, and robust control. Still relatively few works leverage deep Bayesian models for control \citep{gal2016deeppilco,azizzadenesheli2018bdqn}. We argue that this might be because implementing and training these models can be difficult and time consuming. To demonstrate our module, we implement Bayesian PlaNet, based on the work of \citet{hafner2018planet}. The original PlaNet agent learns a latent dynamics model as a sequential VAE on image observations. A sample-based planner then searches for the most promising action sequence in the latent space of the model.

We extend this agent by changing the fully connected layers of the transition function to their Bayesian counterparts, \texttt{VariationalDense}. Swapping the layers and adding the KL term to the loss,  we reach a score of 614 on the cheetah task, matching the performance of the original agent. We monitor the KL divergence of the weight posterior to verify that the model indeed learns a non-trivial belief. This result demonstrates that incorporating model estimates into reinforcement learning agents can be straightforward given the right software abstractions. The fact that the same performance is achieved opens up many possible approaches for exploration and robust control; see \Cref{appendix:velocity}.

\section{Discussion}
\label{sec:limitations}

We described Bayesian Layers, a module designed for fast experimentation with neural network uncertainty. By capturing uncertainty-aware functions, Bayesian Layers lets one naturally experiment with and scale up Bayesian neural networks, GPs, and flow-based models.

\clearpage
\bibliographystyle{apalike}
\bibliography{bib}

\begin{thebibliography}{}

\bibitem[Abadi et~al., 2015]{tensorflow2015-whitepaper}
Abadi, M., Agarwal, A., Barham, P., Brevdo, E., Chen, Z., Citro, C., Corrado,
  G.~S., Davis, A., Dean, J., Devin, M., Ghemawat, S., Goodfellow, I., Harp,
  A., Irving, G., Isard, M., Jia, Y., Jozefowicz, R., Kaiser, L., Kudlur, M.,
  Levenberg, J., Man\'{e}, D., Monga, R., Moore, S., Murray, D., Olah, C.,
  Schuster, M., Shlens, J., Steiner, B., Sutskever, I., Talwar, K., Tucker, P.,
  Vanhoucke, V., Vasudevan, V., Vi\'{e}gas, F., Vinyals, O., Warden, P.,
  Wattenberg, M., Wicke, M., Yu, Y., and Zheng, X. (2015).
\newblock {TensorFlow}: Large-scale machine learning on heterogeneous systems.
\newblock Software available from tensorflow.org.

\bibitem[{Aboleth Developers}, 2017]{aboleth2017aboleth}
{Aboleth Developers} (2017).
\newblock Aboleth.
\newblock \url{https://github.com/data61/aboleth}.

\bibitem[Al-Rfou et~al., 2016]{alrfou2016theano}
Al-Rfou, R., Alain, G., Almahairi, A., Angermueller, C., Bahdanau, D., Ballas,
  N., Bastien, F., Bayer, J., Belikov, A., Belopolsky, A., Bengio, Y.,
  Bergeron, A., Bergstra, J., Bisson, V., {Bleecher Snyder}, J., Bouchard, N.,
  Boulanger-Lewandowski, N., Bouthillier, X., de~Br\'ebisson, A., Breuleux, O.,
  Carrier, P.-L., Cho, K., Chorowski, J., Christiano, P., Cooijmans, T.,
  C\^ot\'e, M.-A., C\^ot\'e, M., Courville, A., Dauphin, Y.~N., Delalleau, O.,
  Demouth, J., Desjardins, G., Dieleman, S., Dinh, L., Ducoffe, M., Dumoulin,
  V., {Ebrahimi Kahou}, S., Erhan, D., Fan, Z., Firat, O., Germain, M., Glorot,
  X., Goodfellow, I., Graham, M., Gulcehre, C., Hamel, P., Harlouchet, I.,
  Heng, J.-P., Hidasi, B., Honari, S., Jain, A., Jean, S., Jia, K., Korobov,
  M., Kulkarni, V., Lamb, A., Lamblin, P., Larsen, E., Laurent, C., Lee, S.,
  Lefrancois, S., Lemieux, S., L\'eonard, N., Lin, Z., Livezey, J.~A., Lorenz,
  C., Lowin, J., Ma, Q., Manzagol, P.-A., Mastropietro, O., McGibbon, R.~T.,
  Memisevic, R., van Merri\"enboer, B., Michalski, V., Mirza, M., Orlandi, A.,
  Pal, C., Pascanu, R., Pezeshki, M., Raffel, C., Renshaw, D., Rocklin, M.,
  Romero, A., Roth, M., Sadowski, P., Salvatier, J., Savard, F., Schl\"uter,
  J., Schulman, J., Schwartz, G., Serban, I.~V., Serdyuk, D., Shabanian, S.,
  Simon, E., Spieckermann, S., Subramanyam, S.~R., Sygnowski, J., Tanguay, J.,
  van Tulder, G., Turian, J., Urban, S., Vincent, P., Visin, F., de~Vries, H.,
  Warde-Farley, D., Webb, D.~J., Willson, M., Xu, K., Xue, L., Yao, L., Zhang,
  S., and Zhang, Y. (2016).
\newblock {Theano: A {Python} framework for fast computation of mathematical
  expressions}.
\newblock {\em arXiv preprint arXiv:1605.02688}.

\bibitem[Al-Shedivat et~al., 2017]{al2017learning}
Al-Shedivat, M., Wilson, A.~G., Saatchi, Y., Hu, Z., and Xing, E.~P. (2017).
\newblock Learning scalable deep kernels with recurrent structure.
\newblock {\em Journal of Machine Learning Research}, 18(1).

\bibitem[Azizzadenesheli et~al., 2018]{azizzadenesheli2018bdqn}
Azizzadenesheli, K., Brunskill, E., and Anandkumar, A. (2018).
\newblock Efficient exploration through bayesian deep q-networks.
\newblock {\em arXiv preprint arXiv:1802.04412}.

\bibitem[Bingham et~al., 2018]{bingham2018pyro}
Bingham, E., Chen, J.~P., Jankowiak, M., Obermeyer, F., Pradhan, N.,
  Karaletsos, T., Singh, R., Szerlip, P., Horsfall, P., and Goodman, N.~D.
  (2018).
\newblock {Pyro: Deep Universal Probabilistic Programming}.
\newblock {\em arXiv preprint arXiv:1810.09538}.

\bibitem[Blundell et~al., 2015]{blundell2015weight}
Blundell, C., Cornebise, J., Kavukcuoglu, K., and Wierstra, D. (2015).
\newblock Weight uncertainty in neural networks.
\newblock In {\em International Conference on Machine Learning}.

\bibitem[Bui et~al., 2016]{bui2016deepgpep}
Bui, T., Hernandez-Lobato, D., Hernandez-Lobato, J., Li, Y., and Turner, R.
  (2016).
\newblock Deep gaussian processes for regression using approximate expectation
  propagation.
\newblock In Balcan, M.~F. and Weinberger, K.~Q., editors, {\em Proceedings of
  The 33rd International Conference on Machine Learning}, volume~48 of {\em
  Proceedings of Machine Learning Research}, pages 1472--1481, New York, New
  York, USA. PMLR.

\bibitem[Carpenter et~al., 2016]{carpenter2016stan}
Carpenter, B., Gelman, A., Hoffman, M.~D., Lee, D., Goodrich, B., Betancourt,
  M., Brubaker, M., Guo, J., Li, P., and Riddell, A. (2016).
\newblock Stan: {A} probabilistic programming language.
\newblock {\em Journal of Statistical Software}.

\bibitem[Chelba et~al., 2013]{chelba2013one}
Chelba, C., Mikolov, T., Schuster, M., Ge, Q., Brants, T., and Koehn, P.
  (2013).
\newblock One billion word benchmark for measuring progress in statistical
  language modeling.
\newblock {\em CoRR}, abs/1312.3005.

\bibitem[Chen et~al., 2015]{chen2015mxnet}
Chen, T., Li, M., Li, Y., Lin, M., Wang, N., Wang, M., Xiao, T., Xu, B., Zhang,
  C., and Zhang, Z. (2015).
\newblock {MXNet}: A flexible and efficient machine learning library for
  heterogeneous distributed systems.
\newblock {\em arXiv preprint arXiv:1512.01274}.

\bibitem[Chollet, 2016]{chollet2015keras}
Chollet, F. (2016).
\newblock Keras.
\newblock \url{https://github.com/fchollet/keras}.

\bibitem[Collobert et~al., 2011]{collobert2011torch}
Collobert, R., Kavukcuoglu, K., and Farabet, C. (2011).
\newblock Torch7: A matlab-like environment for machine learning.
\newblock In {\em BigLearn, NIPS Workshop}.

\bibitem[Damianou and Lawrence, 2013]{damianou2013deep}
Damianou, A. and Lawrence, N. (2013).
\newblock Deep gaussian processes.
\newblock In {\em Artificial Intelligence and Statistics}, pages 207--215.

\bibitem[Dillon et~al., 2017]{dillon2017tensorflow}
Dillon, J.~V., Langmore, I., Tran, D., Brevdo, E., Vasudevan, S., Moore, D.,
  Patton, B., Alemi, A., Hoffman, M., and Saurous, R.~A. (2017).
\newblock {TensorFlow Distributions}.
\newblock {\em arXiv preprint arXiv:1711.10604}.

\bibitem[Dinh et~al., 2017]{dinh2017density}
Dinh, L., Sohl-Dickstein, J., and Bengio, S. (2017).
\newblock Density estimation using real nvp.
\newblock In {\em International Conference on Learning Representations}.

\bibitem[Fortunato et~al., 2017]{fortunato2017bayesian}
Fortunato, M., Blundell, C., and Vinyals, O. (2017).
\newblock Bayesian recurrent neural networks.
\newblock {\em arXiv preprint arXiv:1704.02798}.

\bibitem[Gal and Ghahramani, 2016]{gal2016dropout}
Gal, Y. and Ghahramani, Z. (2016).
\newblock Dropout as a bayesian approximation: Representing model uncertainty
  in deep learning.
\newblock In {\em international conference on machine learning}, pages
  1050--1059.

\bibitem[Gal et~al., 2016]{gal2016deeppilco}
Gal, Y., McAllister, R., and Rasmussen, C.~E. (2016).
\newblock Improving pilco with bayesian neural network dynamics models.
\newblock In {\em Data-Efficient Machine Learning workshop, ICML}, volume~4.

\bibitem[Gardner et~al., 2018]{gardner2018gpytorch}
Gardner, J.~R., Pleiss, G., Bindel, D., Weinberger, K.~Q., and Wilson, A.~G.
  (2018).
\newblock Gpytorch: Blackbox matrix-matrix gaussian process inference with gpu
  acceleration.
\newblock In {\em NeurIPS}.

\bibitem[Germain et~al., 2015]{germain2015made}
Germain, M., Gregor, K., Murray, I., and Larochelle, H. (2015).
\newblock Made: Masked autoencoder for distribution estimation.
\newblock In {\em International Conference on Machine Learning}, pages
  881--889.

\bibitem[Gomez et~al., 2017]{gomez2017reversible}
Gomez, A.~N., Ren, M., Urtasun, R., and Grosse, R.~B. (2017).
\newblock The reversible residual network: Backpropagation without storing
  activations.
\newblock In {\em Neural Information Processing Systems}.

\bibitem[Goodman et~al., 2012]{goodman2012church}
Goodman, N., Mansinghka, V., Roy, D.~M., Bonawitz, K., and Tenenbaum, J.~B.
  (2012).
\newblock Church: a language for generative models.
\newblock {\em arXiv preprint arXiv:1206.3255}.

\bibitem[{GPy}, 2012]{gpy2014}
{GPy} (since 2012).
\newblock {GPy}: A gaussian process framework in python.
\newblock \url{http://github.com/SheffieldML/GPy}.

\bibitem[Hafner et~al., 2018a]{hafner2018planet}
Hafner, D., Lillicrap, T., Fischer, I., Villegas, R., Ha, D., Lee, H., and
  Davidson, J. (2018a).
\newblock Learning latent dynamics for planning from pixels.
\newblock {\em arXiv preprint arXiv:1811.04551}.

\bibitem[Hafner et~al., 2018b]{hafner2018reliable}
Hafner, D., Tran, D., Irpan, A., Lillicrap, T., and Davidson, J. (2018b).
\newblock Reliable uncertainty estimates in deep neural networks using noise
  contrastive priors.
\newblock {\em arXiv preprint}.

\bibitem[Hensman et~al., 2013]{hensman2013svi}
Hensman, J., Fusi, N., and Lawrence, N.~D. (2013).
\newblock Gaussian processes for big data.
\newblock In {\em Conference on Uncertainty in Artificial Intelligence}.

\bibitem[Hern\'{a}ndez-Lobato and Adams, 2015]{miguel2015pbp}
Hern\'{a}ndez-Lobato, J.~M. and Adams, R.~P. (2015).
\newblock Probabilistic backpropagation for scalable learning of bayesian
  neural networks.
\newblock In {\em Proceedings of the 32Nd International Conference on
  International Conference on Machine Learning - Volume 37}, ICML'15, pages
  1861--1869. JMLR.org.

\bibitem[Jia et~al., 2014]{jia2014caffe}
Jia, Y., Shelhamer, E., Donahue, J., Karayev, S., Long, J., Girshick, R.,
  Guadarrama, S., and Darrell, T. (2014).
\newblock Caffe: Convolutional architecture for fast feature embedding.
\newblock In {\em Proceedings of the 22nd ACM international conference on
  Multimedia}, pages 675--678. ACM.

\bibitem[John and Hensman, 2018]{john2018large}
John, S.~T. and Hensman, J. (2018).
\newblock Large-scale cox process inference using variational fourier features.
\newblock {\em arXiv preprint arXiv:1804.01016}.

\bibitem[Jouppi et~al., 2017]{jouppi2017datacenter}
Jouppi, N.~P., Young, C., Patil, N., Patterson, D., Agrawal, G., Bajwa, R.,
  Bates, S., Bhatia, S., Boden, N., Borchers, A., et~al. (2017).
\newblock In-datacenter performance analysis of a tensor processing unit.
\newblock In {\em Proceedings of the 44th Annual International Symposium on
  Computer Architecture}.

\bibitem[Kingma and Welling, 2014]{kingma2014auto}
Kingma, D.~P. and Welling, M. (2014).
\newblock Auto-encoding variational {B}ayes.
\newblock In {\em International Conference on Learning Representations}.

\bibitem[Kiselyov and Shan, 2009]{kiselyov2009embedded}
Kiselyov, O. and Shan, C.-C. (2009).
\newblock Embedded probabilistic programming.
\newblock In {\em DSL}, volume 5658, pages 360--384. Springer.

\bibitem[Laumann and Shridhar, 2018]{laumann2018bayesian}
Laumann, F. and Shridhar, K. (2018).
\newblock Bayesian convolutional neural networks.
\newblock {\em arXiv preprint arXiv:1806.05978}.

\bibitem[Louizos and Welling, 2017]{louizos2017multiplicative}
Louizos, C. and Welling, M. (2017).
\newblock Multiplicative normalizing flows for variational bayesian neural
  networks.
\newblock {\em arXiv preprint arXiv:1703.01961}.

\bibitem[Matthews et~al., 2017]{GPflow2017}
Matthews, A. G. d.~G., {van der Wilk}, M., Nickson, T., Fujii, K.,
  {Boukouvalas}, A., {Le{\'o}n-Villagr{\'a}}, P., Ghahramani, Z., and Hensman,
  J. (2017).
\newblock {{GP}flow: A {G}aussian process library using {T}ensor{F}low}.
\newblock {\em Journal of Machine Learning Research}, 18(40):1--6.

\bibitem[Narayanan et~al., 2016]{narayanan2016probabilistic}
Narayanan, P., Carette, J., Romano, W., Shan, C.-c., and Zinkov, R. (2016).
\newblock {Probabilistic Inference by Program Transformation in Hakaru (System
  Description)}.
\newblock In {\em International Symposium on Functional and Logic Programming},
  pages 62--79, Cham. Springer, Cham.

\bibitem[Neal, 1995]{neal1995software}
Neal, R. (1995).
\newblock Software for flexible bayesian modeling and markov chain sampling.
\newblock \url{https://www.cs.toronto.edu/~radford/fbm.software.html}.

\bibitem[Parmar et~al., 2018]{parmar2018image}
Parmar, N., Vaswani, A., Uszkoreit, J., Kaiser, {\L}., Shazeer, N., Ku, A., and
  Tran, D. (2018).
\newblock Image transformer.
\newblock In {\em International Conference on Machine Learning}.

\bibitem[{Probtorch Developers}, 2017]{probtorch2017probtorch}
{Probtorch Developers} (2017).
\newblock Probtorch.
\newblock \url{https://github.com/probtorch/probtorch}.

\bibitem[Rasmussen and Nickisch, 2010]{rasmussen2010gaussian}
Rasmussen, C.~E. and Nickisch, H. (2010).
\newblock Gaussian processes for machine learning (gpml) toolbox.
\newblock {\em Journal of machine learning research}, 11(Nov):3011--3015.

\bibitem[Rezende and Mohamed, 2015]{rezende2015variational}
Rezende, D.~J. and Mohamed, S. (2015).
\newblock Variational inference with normalizing flows.
\newblock In {\em International Conference on Machine Learning}.

\bibitem[S. and N., 2016]{guadarrama2016slim}
S., G. and N., S. (2016).
\newblock {TensorFlow-Slim}: A lightweight library for defining, training and
  evaluating complex models in {TensorFlow}.

\bibitem[Salimans et~al., 2017]{salimans2017pixelcnnpp}
Salimans, T., Karpathy, A., Chen, X., and Kingma, D.~P. (2017).
\newblock {PixelCNN++}: Improving the pixelcnn with discretized logistic
  mixture likelihood and other modifications.
\newblock {\em arXiv preprint arXiv:1701.05517}.

\bibitem[Salimbeni and Deisenroth, 2017]{salimbeni2017doubly}
Salimbeni, H. and Deisenroth, M. (2017).
\newblock Doubly stochastic variational inference for deep gaussian processes.
\newblock In {\em Advances in Neural Information Processing Systems}, pages
  4588--4599.

\bibitem[Shazeer et~al., 2018]{shazeer2018mesh}
Shazeer, N., Cheng, Y., Parmar, N., Tran, D., Vaswani, A., Koanantakool, P.,
  Hawkins, P., Lee, H., Hong, M., Young, C., Sepassi, R., and Hechtman, B.
  (2018).
\newblock {Mesh-TensorFlow}: Deep learning for supercomputers.
\newblock In {\em Neural Information Processing Systems}.

\bibitem[Tran et~al., 2018]{tran2018simple}
Tran, D., Hoffman, M.~D., Moore, D., Suter, C., Vasudevan, S., Radul, A.,
  Johnson, M., and Saurous, R.~A. (2018).
\newblock Simple, distributed, and accelerated probabilistic programming.
\newblock In {\em Neural Information Processing Systems}.

\bibitem[Tran et~al., 2016]{tran2016edward}
Tran, D., Kucukelbir, A., Dieng, A.~B., Rudolph, M., Liang, D., and Blei, D.~M.
  (2016).
\newblock {Edward: A library for probabilistic modeling, inference, and
  criticism}.
\newblock {\em arXiv preprint arXiv:1610.09787}.

\bibitem[Vanhatalo et~al., 2013]{vanhatalo2013gpstuff}
Vanhatalo, J., Riihim{\"a}ki, J., Hartikainen, J., Jyl{\"a}nki, P., Tolvanen,
  V., and Vehtari, A. (2013).
\newblock Gpstuff: Bayesian modeling with gaussian processes.
\newblock {\em Journal of Machine Learning Research}, 14(Apr):1175--1179.

\bibitem[Vaswani et~al., 2018]{vaswani2018tensor2tensor}
Vaswani, A., Bengio, S., Brevdo, E., Chollet, F., Gomez, A.~N., Gouws, S.,
  Jones, L., Kaiser, L., Kalchbrenner, N., Parmar, N., Sepassi, R., Shazeer,
  N., and Uszkoreit, J. (2018).
\newblock Tensor2tensor for neural machine translation.
\newblock {\em CoRR}, abs/1803.07416.

\bibitem[Wen et~al., 2018]{wen2018flipout}
Wen, Y., Vicol, P., Ba, J., Tran, D., and Grosse, R. (2018).
\newblock Flipout: Efficient pseudo-independent weight perturbations on
  mini-batches.
\newblock In {\em International Conference on Learning Representations}.

\end{thebibliography}
\clearpage

\appendix
\onecolumn

\section{Bayesian ResNet-50}
\label{appendix:cnn}

See \Cref{fig:cnn}.

\newenvironment{widespace}[1]
{\dimen0=#1\relax
 \dimen1=\dimexpr 0.5\dimen0 - 0.5\textwidth\relax
 \hspace*{-\dimen1}\minipage{\dimen0}}%
{\endminipage\hspace{-\dimen1}}

\begin{figure*}[!htbp]

\begin{widespace}{7in}
\begin{minipage}{.5\textwidth}
  \input{_minted-main/4CCCF35B5CC1DA390CE7533CD733135EB722149D8701C8F8F9BC8FCEDA70B7B8.pygtex}

\end{minipage}\hfill
\begin{minipage}{.48\textwidth}
  \input{_minted-main/2EA23380C24563B86F636D93F2AE4D04B722149D8701C8F8F9BC8FCEDA70B7B8.pygtex}
\end{minipage}

\vspace{3ex}

\begin{minipage}{.15\textwidth}\hfill\end{minipage}\hfill  
\begin{minipage}{.85\textwidth}
  \input{_minted-main/A72A9D17E50027239408B651BF36F2BEB722149D8701C8F8F9BC8FCEDA70B7B8.pygtex}

\end{minipage}
\end{widespace}
\caption{Bayesian ResNet-50.}
\label{fig:cnn}
\end{figure*}

\section{Bayesian PlaNet}
\label{appendix:velocity}

\begin{figure}[!htb]
\centering
\begin{minipage}{.3\textwidth}
\includegraphics[width=\textwidth]{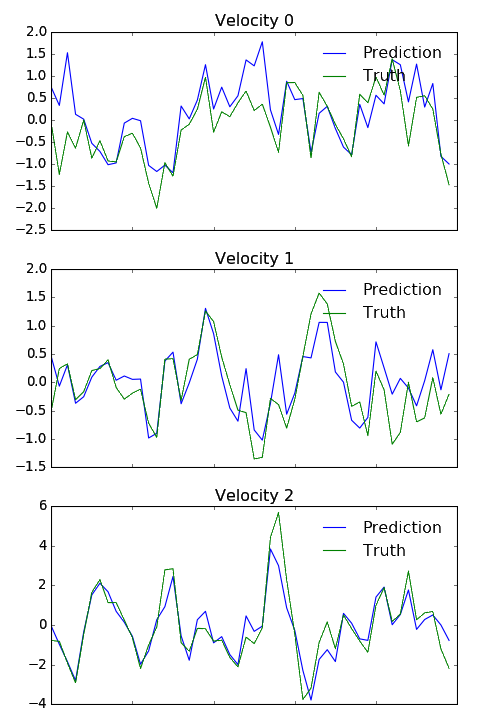}
\end{minipage}\hfill
\begin{minipage}{.3\textwidth}
\includegraphics[width=\textwidth]{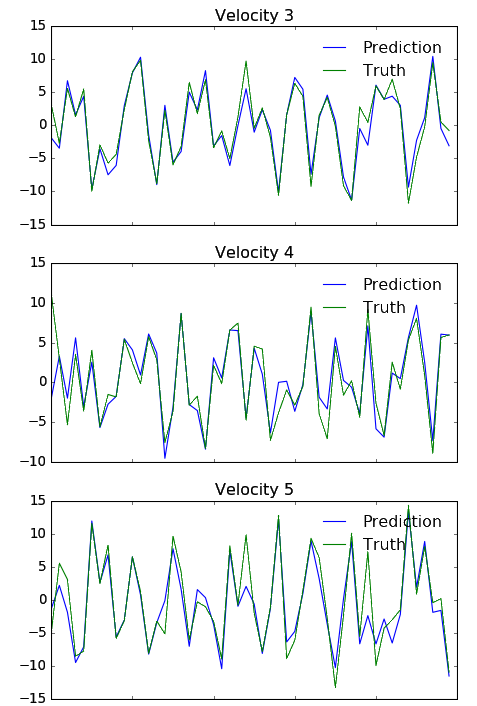}
\end{minipage}\hfill
\begin{minipage}{.3\textwidth}
\includegraphics[width=\textwidth]{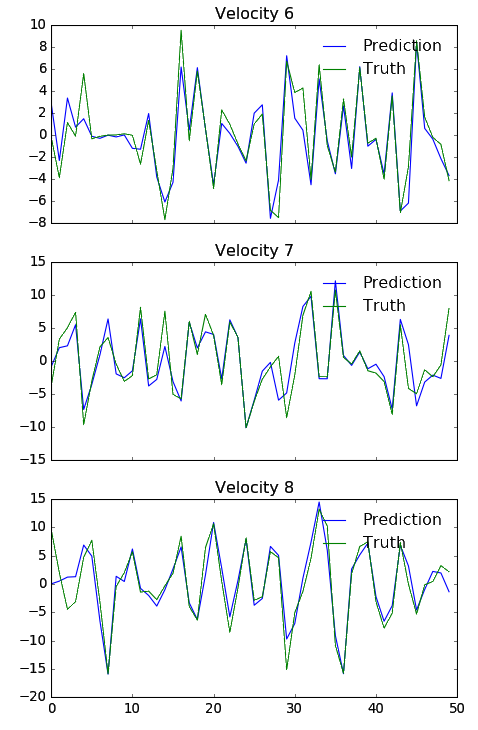}
\end{minipage}
\caption{Given the Bayesian PlaNet agent, we predict the true velocities of the reinforcement learning environment from its encoded latent states. Compared to Figure 7 of \citet{hafner2018planet}, Bayesian PlaNet appears to capture more information in the latent codes resulting in more precise velocity predictions (``world knowledge'').}
\label{fig:velocity}
\end{figure}

See \Cref{fig:velocity}.

\end{document}